\documentclass{sig-alternate}
\usepackage{color}
\usepackage{paralist}
\usepackage{soul}
\usepackage{url}
\usepackage[boxed, linesnumbered]{algorithm2e}
\makeatletter
\def\@copyrightspace{\relax}
\makeatother
\usepackage[
  pass,
]{geometry}
\soulregister\cite7
\soulregister\ref7

\begin{document}
%
\numberofauthors{1} 
%
\author{
%
%
\alignauthor
Zuxuan Wu, Xi Wang, Yu-Gang Jiang\titlenote{Corresponding author.}, Hao Ye, Xiangyang Xue\\
\affaddr{School of Computer Science, Shanghai Key Lab of Intelligent Information 
Processing, \\ Fudan University, Shanghai, China}
\email{\{zxwu, xwang10, ygj, haoye10, xyxue\}@fudan.edu.cn}
}


\title{Modeling Spatial-Temporal Clues in a Hybrid Deep Learning Framework for Video Classification}
\maketitle
\begin{abstract}
Classifying videos according to content semantics is an important problem with a wide range of applications. 
In this paper, we propose a hybrid deep learning framework for video classification, which is able to model static spatial information, short-term motion, as well as long-term temporal clues in the videos. Specifically, the spatial and the short-term motion features are extracted separately by two Convolutional Neural Networks (CNN). These two types of CNN-based features are then combined in a regularized feature fusion network for classification, which is able to learn and utilize feature relationships for improved performance. In addition, Long Short Term Memory (LSTM) networks are applied on top of the two features to further model longer-term temporal clues. The main contribution of this work is the hybrid learning framework that can model several important aspects of the video data. We also show that (1) combining the spatial and the short-term motion features in the regularized fusion network is better than direct classification and fusion using the CNN with a softmax layer, and (2) the sequence-based LSTM is highly complementary to the traditional classification strategy without considering the temporal frame orders. Extensive experiments are conducted on two popular and challenging benchmarks, the UCF-101 Human Actions and the Columbia Consumer Videos (CCV). On both benchmarks, our framework achieves to-date the best reported performance: $91.3\%$ on the UCF-101 and $83.5\%$ on the CCV.
\end{abstract}

\vspace{0.2in}

\section{Introduction}
Video classification based on contents like human actions or complex events is a challenging task that has been extensively studied in the research community. Significant progress has been achieved in recent years by designing various features, which are expected to be robust to intra-class variations and discriminative to separate different classes. For example, one can utilize traditional image-based features like the SIFT \cite{lowe2004distinctive} to capture the static spatial information in videos. In addition to the static frame based visual features, motion is a very important clue for video classification, as most classes containing object movements like the human actions require the motion information to be reliably recognized. For this, a very popular feature is the dense trajectories~\cite{wang2013action}, which tracks densely sampled local frame patches over time and computes several traditional features based on the trajectories.

In contrast to the hand-crafted features, there is a growing trend of learning robust feature representations from raw data with deep neural networks. Among the many existing network structures, Convolutional Neural Networks (CNN) have demonstrated great success on various tasks, including image classification~\cite{krizhevsky2012imagenet,Szegedy:2014tb,simonyan2014very}, image-based object localization~\cite{girshick2014rcnn}, speech recognition~\cite{Dahl:2012dx}, \textit{etc}. For video classification, Ji \textit{et al.}~\cite{DBLP:conf/icml/JiXYY10} and Karparthy~\textit{et al.}~\cite{KarpathyCVPR14} extended the CNN to work on the temporal dimension by stacking frames over time. Recently, Simonyan~\textit{et al.}~\cite{DBLP:conf/nips/SimonyanZ14} proposed a two-stream CNN approach, which uses two CNNs on static frames and optical flows respectively to capture the spatial and the motion information. It focuses only on short-term motion as the optical flows are computed in very short time windows. With this approach, similar or slightly better performance than the hand-crafted features like~\cite{wang2013action} has been reported.

These existing works, however, are not able to model the long-term temporal clues in the videos. As aforementioned, the two-stream CNN~\cite{DBLP:conf/nips/SimonyanZ14} uses stacked optical flows computed in short time windows as inputs, and the order of the optical flows is fully discarded in the learning process (cf. Section 3.1). This is not sufficient for video classification, as many complex contents can be better identified by considering the temporal order of short-term actions. Take ``birthday'' event as an example---it usually involves several sequential actions, such as ``making a wish'', ``blowing out candles'' and ``eating cakes''.

\begin{figure*}
\centering
\epsfig{file=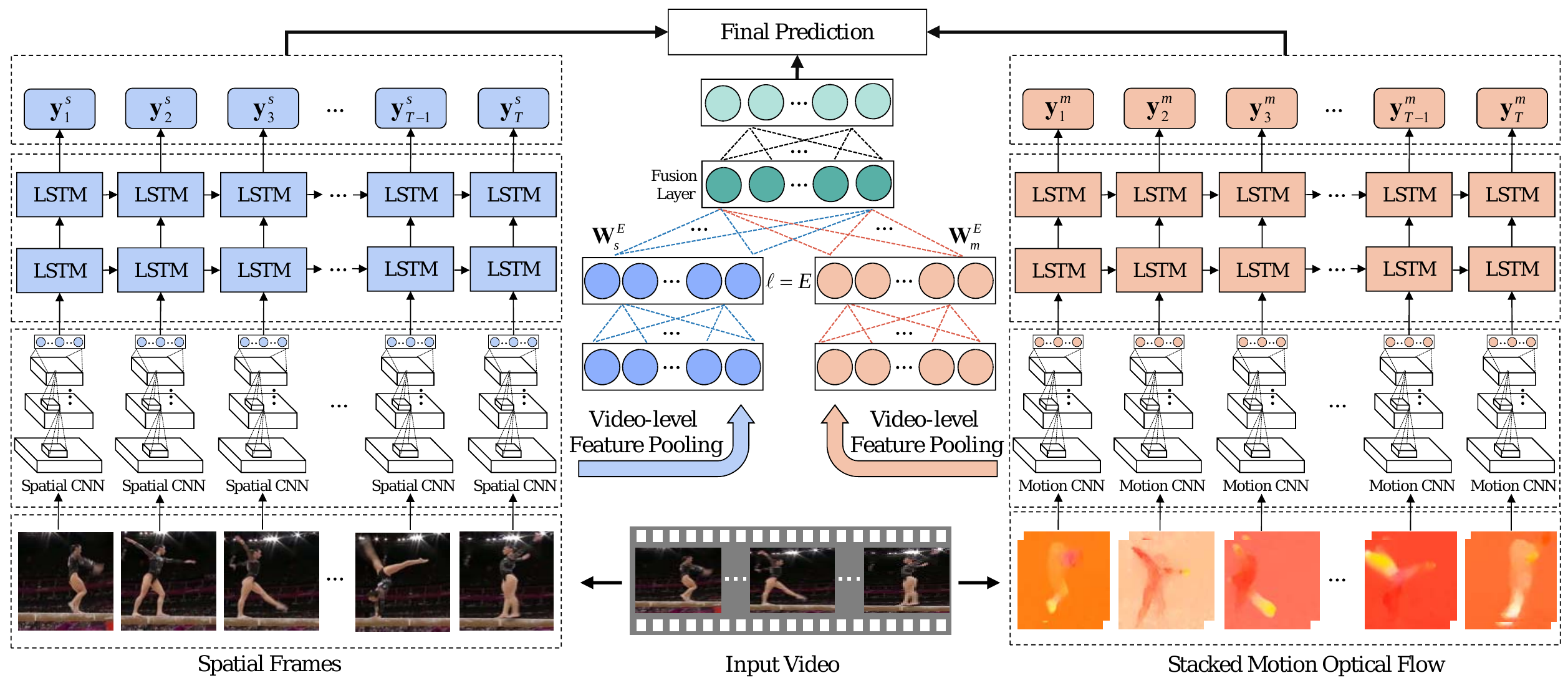, scale=0.695}
\caption{\label{fig:framework}An overview of the proposed hybrid deep learning framework for video classification. Given an input video, two types of features are extracted using the CNN from spatial frames and short-term stacked motion optical flows respectively. The features are separately fed into two sets of LSTM networks for long-term temporal modeling (Left and Right). In addition, we also employ a regularized feature fusion network to perform video-level feature fusion and classification (Middle). The outputs of the sequence-based LSTM and the video-level feature fusion network are combined to generate the final prediction. See texts for more discussions. }
\end{figure*}

To address the above limitation, this paper proposes a hybrid deep learning framework for video classification, which is able to harness not only the spatial and short-term motion features, but also the long-term temporal clues. In order to leverage the temporal information, we adopt a Recurrent Neural Networks (RNN) model called Long Short Term Memory (LSTM), which maps the input sequences to outputs using a sequence of hidden states and incorporates memory units that enable the network to learn when to forget previous hidden states and when to update hidden states with new information. In addition, many approaches fuse multiple features in a very ``shallow" manner by either concatenating the features before classification or averaging the predictions of classifiers trained using different features separately. In this work we integrate the spatial and the short-term motion features in a deep neural network with carefully designed regularizations to explore feature correlations. This method can perform video classification within the same network and further combining its outputs with the predictions of the LSTMs can lead to very competitive classification performance.

Figure~\ref{fig:framework} gives an overview of the proposed framework. Spatial and short-term motion features are first extracted by the two-stream CNN approach~\cite{DBLP:conf/nips/SimonyanZ14}, and then input into the LSTM for long-term temporal modeling. Average pooling is adopted to generate video-level spatial and motion features, which are fused by the regularized feature fusion network. After that, outputs of the sequence-based LSTM and the video-level feature fusion network are combined as the final predictions. Notice that, in contrast to the current framework, alternatively one may train a fusion network to combine the frame-level spatial and motion features first and then use a single set of LSTM for temporal modeling. However, in our experiments we have observed worse results using this strategy. The main reason is that learning dimension-wise feature correlations in the fusion network requires strong and reliable supervision, but we only have video-level class labels which are not necessarily always related to the frame semantics. In other words, the imprecise frame-level labels populated from the video annotations are too noisy to learn a good fusion network. The main contributions of this work are summarized as follows:
\begin{itemize}
\item We propose an end-to-end hybrid deep learning framework for video classification, which can model not only the short-term spatial-motion patterns but also the long-term temporal clues with variable-length video sequences as inputs. 

\item We adopt the LSTM to model long-term temporal clues on top of both the spatial and the short-term motion features. We show that both features work well with the LSTM, and the LSTM based classifiers are very complementary to the traditional classifiers without considering the temporal frame orders. 

\item We fuse the spatial and the motion features in a regularized feature fusion network that can explore feature correlations and perform classification. The network is computationally efficient in both training and testing.

\item Through an extensive set of experiments, we demonstrate that our proposed framework outperforms several alternative methods with clear margins. On the well-known UCF-101 and CCV benchmarks, we attain to-date the best performance.
\end{itemize}

The rest of this paper is organized as follows. Section~2 reviews related works. Section~3 describes the proposed hybrid deep learning framework in detail. Experimental results and comparisons are discussed in Section~4, followed by conclusions in Section~5.
\vspace{0.2in}
\section{Related Works}
Video classification has been a longstanding research topic in multimedia and computer vision. Successful classification systems rely heavily on the extracted video features, and hence most existing works focused on designing robust and discriminative features. Many video representations were motivated by the advances in image domain, which can be extended to utilize the temporal dimension of the video data. For instance, Laptev~\cite{laptevSTIP} extended the 2D Harris corner detector \cite{Harris1988} into 3D space to find space-time interest points. Klaser \textit{et al.} proposed HOG3D by extending the idea of integral images for fast descriptor computation~\cite{klaser2008spatio}. Wang \textit{et al.} reported that dense sampling at regular positions in space and time outperforms the detected sparse interest points on video classification tasks~\cite{wang2009evaluation}. Partly inspired by this finding, they further proposed the dense trajectory features, which densely sample local patches from each frame at different scales and then track them in a dense optical flow field over time~\cite{wang2013action}. This method has demonstrated very competitive results on major benchmark datasets. In addition, further improvements may be achieved by using advantageous feature encoding methods like the Fisher Vectors~\cite{oneata2013action} or adopting feature normalization strategies, such as RootSift~\cite{arandjelovic2012three} and Power Norm~\cite{sanchez2013image}. Note that these spatial-temporal video descriptors only capture local motion patterns within a very short period, and popular descriptor quantization methods like the bag-of-words entirely destroy the temporal order information of the descriptors. 

To explore the long-term temporal clues, graphical models have been popularly used, such as hidden Markov models (HMM), Bayesian Networks (BN), Conditional Random Fields (CRF), \textit{etc}. For instance, Li \textit{et al.} proposed to replace the hidden states in HMMs with visualizable salient poses estimated by Gaussian Mixture Models~\cite{li2008expandable}, and Tang~\textit{et al.} introduced latent variables over video frames to discover the most discriminative states of an event based on a variable duration HMM~\cite{tang2012learning}. Zeng \textit{et al.} exploited multiple types of domain knowledge to guide the learning of a Dynamic BN for action recognition~\cite{zeng2010knowledge}. Instead of using directed graphical models like the HMM and BN, undirected graphical models have also been adopted. Vail \textit{et al.} employed the CRF for activity recognition in~\cite{vail2007conditional}. Wang \textit{et al.} proposed a max-margin hidden CRF for action recognition in videos, where a human action is modeled as a global root template and a constellation of several ``parts''~\cite{wang2009max}. 

Many related works have investigated the fusion of multiple features, which is often effective for improving classification performance. The most straightforward and popular ways are early fusion and late fusion. Generally, the early fusion refers to fusion at the feature level, such as feature concatenation or linear combination of kernels of individual features. For example, in \cite{zhang2007local}, Zhang \textit{et al.} computed non-linear kernels for each feature separately, and then fused the kernels for model training. The fusion weights can be manually set or automatically estimated by multiple kernel learning (MKL)~\cite{bach2004multiple}. For the late fusion methods, independent classifiers are first trained using each feature separately, and outputs of the classifiers are then combined. In \cite{ye2012robust}, Ye~\textit{et al.} proposed a robust late fusion approach to fuse multiple classification outputs by seeking a shared low-rank latent matrix, assuming that noises may exist in the predictions of some classifiers, which can possibly be removed by using the low-rank matrix.

Both early and late fusion fail to explore the correlations shared by the features and hence are not ideal for video classification. In this paper we employ a regularized neural network tailored feature fusion and classification, which can automatically learn dimension-wise feature correlations. Several studies are related. In~\cite{jiang2009short,MVA:audiovisual}, the authors proposed to construct an audio-visual joint codebook for video classification, in order to discover and model the audio-visual feature correlations. There are also studies on using neural networks for feature fusion. In~\cite{srivastava2012multimodal}, the authors employed deep Boltzmann machines to learn a fused representation of images and texts. In~\cite{ngiam2011multimodal}, a deep denoised auto-encoder was used for cross-modality and shared representation learning. Very recently, Wu \textit{et al.}~\cite{mm14:videoclassification} presented an approach using regularizations in neural networks to exploit feature and class relationships. The fusion approach in this work differs in the following. First, instead of using the traditional hand-crafted features as inputs, we adopt CNN features trained from both static frames and motion optical flows. Second and very importantly, the formulation in the regularized feature fusion network has a much lower complexity compared with that of~\cite{mm14:videoclassification}. 

Researchers have attempted to apply the RNN to model the long-term temporal information in videos. Venugopalan \textit{et al.}~\cite{DBLP:journals/corr/VenugopalanXDRMS14} proposed to translate videos to textual sentences with the LSTM through transferring knowledge from image description tasks. Ranzato~\textit{et al.}~\cite{DBLP:journals/corr/RanzatoSBMCC14} introduced a generative model with the RNN to predict motions in videos. In the context of video classification, Donahua~\textit{et al.} adopted the LSTM to model temporal information~\cite{DBLP:journals/corr/DonahueHGRVSD14}
and Srivastava~\textit{et al.} designed an encoder-decoder RNN architecture to learn feature representations in an unsupervised manner~\cite{DBLP:journals/corr/SrivastavaMS15}. To model motion information, both works adopted optical flow ``images" between nearby frames as the inputs of the LSTM. In contrast, our approach adopts stacked optical flows. Stacked flows over a short time period can better reflect local motion patterns, which are found to be able to produce better results. In addition, our framework incorporates video-level predictions with the feature fusion network for significantly improved performance, which was not considered in these existing works. 

Besides the above discussions of related studies on feature fusion and temporal modeling with the RNN, several representative CNN-based approaches for video classification should also be covered here. The image-based CNN features have recently been directly adopted for video classification, extracted using off-the-shelf models trained on large-scale image datasets like the ImageNet~\cite{JainTHUMOS14,DBLP:journals/corr/RazavianASC14,zha2015exploiting}. For instance, Jain~\textit{et al.}~\cite{JainTHUMOS14} performed action recognition using the SVM classifier with such CNN features and achieved top results in the 2014 THUMOS action recognition challenge~\cite{THUMOS14}. A few works have also tried to extend the CNN to exploit the motion information in videos. Ji~\textit{et al.}~\cite{DBLP:conf/icml/JiXYY10} and Karparthy \textit{et al.}~\cite{KarpathyCVPR14} extended the CNN by stacking visual frames in fixed-size time windows and using spatial-temporal convolutions for video classification. Differently, the two-stream CNN approach by Simonyan~\textit{et al.}~\cite{DBLP:conf/nips/SimonyanZ14} applies the CNN separately on visual frames (the spatial stream) and stacked optical flows (the motion stream). This approach has been found to be more effective, which is adopted as the basis of our proposed framework. However, as discussed in Section~1, all these approaches~\cite{DBLP:conf/icml/JiXYY10,KarpathyCVPR14,DBLP:conf/nips/SimonyanZ14} can only model short-term motion, not the long-term temporal clues.

\section{Methodology}
In this section, we describe the key components of the proposed hybrid deep learning framework shown in Figure~\ref{fig:framework}, including the CNN-based spatial and short-term motion features, the long-term LSTM-based temporal modeling, and the video-level regularized feature fusion network.

\subsection{Spatial and Motion CNN Features}
\begin{figure}[t!]
\centering
\includegraphics[scale=1]{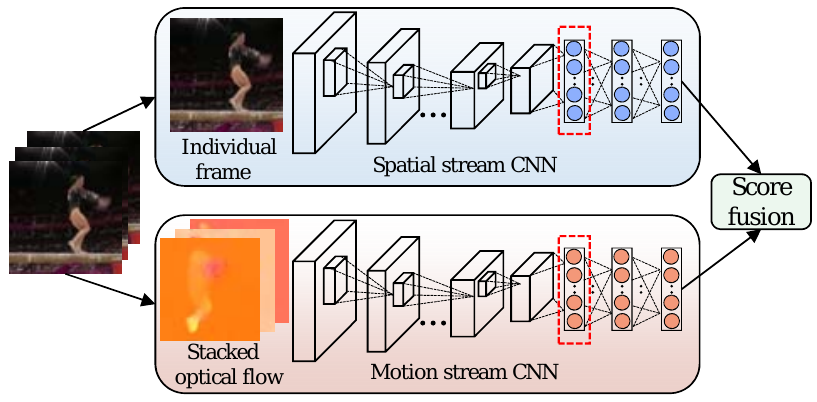}
\caption{The framework of the two-stream CNN. Outputs of the first fully-connected layer in the two CNNs (outlined) are used as the spatial and the short-term motion features for further processing. }
\label{fig:twostream}
\end{figure}

Conventional CNN architectures take images as the inputs and consist of alternating convolutional and pooling layers, which are further topped by a few fully-connected (FC) layers. To extract the spatial and the short-term motion features, we adopt the recent two-stream CNN approach~\cite{DBLP:conf/nips/SimonyanZ14}. Instead of using stacked frames in short time windows like~\cite{DBLP:conf/icml/JiXYY10,KarpathyCVPR14}, this approach decouples the videos into spatial and motion\footnote{\small Note that the authors of the original paper \cite{DBLP:conf/nips/SimonyanZ14} used the name ``temporal stream". We call it ``motion stream" as it only captures short-term motion, which is different from the long-term temporal modeling in our proposed framework. } streams modeled by two CNNs separately. Figure~\ref{fig:twostream} gives an overview. The spatial stream is built on sampled individual frames, which is exactly the same as the CNN-based image classification pipeline and is suitable for capturing the static information in videos like scene backgrounds and basic objects. The motion counterpart operates on top of stacked optical flows. Specifically, optical flows (displacement vector fields) are computed between each pair of adjacent frames, and the horizontal and vertical components of the displacement vectors can form two optical flow images. Instead of using each individual flow image as the input of the CNN, it was reported that stacked optical flows over a time window are better due to the ability of modeling the short-term motion. In other words, the input of the motion stream CNN is a $2L$-channel stacked optical flow image, where $L$ is the number of frames in the window.  The two CNNs produce classification scores separately using a softmax layer and the scores are linearly combined as the final prediction. 

Like many existing works on visual classification using the CNN features~\cite{DBLP:journals/corr/RazavianASC14}, we adopt the output of the first FC layer of the two CNNs as the spatial and the short-term motion features.

\subsection{Temporal Modeling with LSTM}
During the training process of the spatial and the motion stream CNNs, each sweep through the network takes one visual frame or one stacked optical flow image, and the temporal order of the frames is fully discarded. To model the long-term dynamic information in video sequences, we leverage the LSTM model, which has been successfully applied to speech recognition~\cite{DBLP:conf/icassp/GravesMH13}, image captioning~\cite{DBLP:journals/corr/DonahueHGRVSD14}, \textit{etc}. LSTM is a type of RNN with controllable memory units and is effective in many long range sequential modeling tasks without suffering from the ``vanishing gradients'' effect like traditional RNNs. Generally, LSTM recursively maps the input representations at the current time step to output labels via a sequence of hidden states, and thus the learning process of LSTM should be in a sequential manner (from left to right in the two sets of LSTM of Figure~\ref{fig:framework}). Finally, we can obtain a prediction score at each time step with a softmax transformation using the hidden states from the last layer of the LSTM.

\begin{figure}[t]
\centering
\epsfig{file=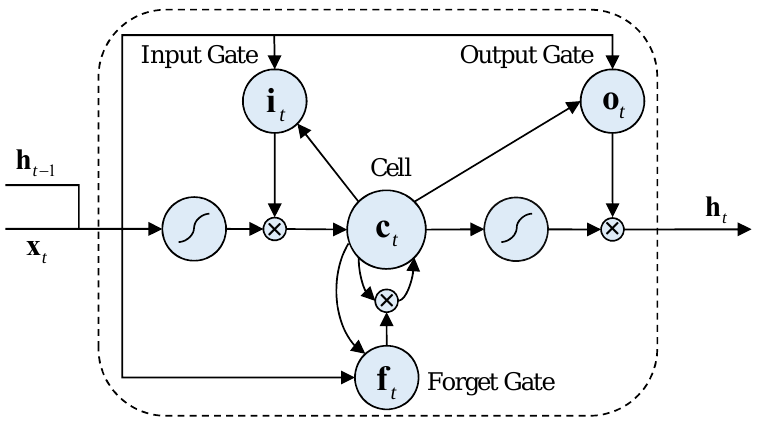, scale=1}
\caption{\label{fig:lstmunit}The structure of  an LSTM unit.}
\end{figure}

More formally, given a sequence of feature representations $({\bf x}_1,{\bf x}_2,\ldots,{\bf x}_T)$, an LSTM maps the inputs to an output sequence  $({\bf y}_1,{\bf y}_2,\ldots,{\bf y}_T)$ by computing activations of the units in the network with the following equations recursively from $t=1$ to $t = T$:
\begin{align*} 
& {\bf i}_t=\sigma({\bf W}_{xi}{\bf x}_t+{\bf W}_{hi}{\bf h}_{t-1}+{\bf W}_{ci}{\bf c}_{t-1}+{\bf b}_i), \\ 
& {\bf f}_t=\sigma({\bf W}_{xf}{\bf x}_t+{\bf W}_{hf}{\bf h}_{t-1}+{\bf W}_{cf}{\bf c}_{t-1}+{\bf b}_f), \\
& {\bf c}_t={\bf f}_t{\bf c}_{t-1}+{\bf i}_t\tanh({\bf W}_{xc}{\bf x}_t+{\bf W}_{hc}{\bf h}_{t-1}+{\bf b}_c), \\
& {\bf o}_t=\sigma({\bf W}_{xo}{\bf x}_t+{\bf W}_{ho}{\bf h}_{t-1}+{\bf W}_{co}{\bf c}_{t}+{\bf b_o}), \\
& {\bf h}_t={\bf o}_t\tanh({\bf c}_t),
\end{align*}
where ${\bf x}_t, {\bf h}_t$ are the input and hidden vectors with the subscription $t$ denoting the $t$-th time step, ${\bf i}_t, {\bf f}_t, {\bf c}_t, {\bf o}_t$ are respectively the activation vectors of the input gate, forget gate, memory cell and output gate, ${\bf W}_{\alpha\beta}$ is the weight matrix between $\alpha$ and $\beta$ (e.g., ${\bf W}_{xi}$ is weight matrix from the input ${\bf x}_t$ to the input gate ${\bf i}_t$), ${\bf b}_\alpha$ is the bias term of $\alpha$ and $\sigma$ is the sigmoid function defined as $\sigma(x) = \frac{1}{1+e^{-x}}$. Figure~\ref{fig:lstmunit} visualizes the structure of an LSTM unit.

The core idea behind the LSTM model is a built-in memory cell that stores information over time to explore long-range dynamics, with non-linear gate units governing the information flow into and out of the cell. As we can see from the above equations, the current frame ${\bf x}_t$ and the previous hidden states ${\bf h}_{t-1}$ are used as inputs of four parts at the $t$-th time step. The memory cell aggregates information from two sources: the previous cell memory unit ${\bf c}_{t-1}$ multiplied by the activation of the forget gate ${\bf f}_t$ and the squashed inputs regulated with the input gate's activation ${\bf i}_t$, the combination of which enables the LSTM to learn to forget information from previous states or consider new information. In addition, the output gate ${\bf o}_t$ controls how much information from the memory cell is passed to the hidden states ${\bf h}_t$ for the following time step. With the explicitly controllable memory units and different functional gates, LSTM can explore long-range temporal clues with variable-length inputs. As a neural network, the LSTM model can be easily deepened by stacking the hidden states from a layer $l-1$ as inputs of the next layer $l$.

Let us now consider a model of $K$ layers, the feature vector ${\bf x}_t$ at the $t$-th time step is fed into the first layer of the LSTM together with the hidden state ${\bf h}^1_{t-1}$ in the same layer obtained from the last time step to produce an updated ${\bf h}^1_t$, which will then be used as the inputs of the following layer. Denote $f_W$ as the mapping function from the inputs to the hidden states, the transition from layer $l-1$ to layer $l$ can be written as:
\begin{equation*}
    {\bf h}^l_t = \left\{
                  \begin{array}{ll}
                        f_W({\bf h}_t^{l-1},{\bf h}_{t-1}^l), & l>1; \\
                    f_W({\bf x}_t,{\bf h}_{t-1}^l), & l=1.
                  \end{array}
                \right.
\end{equation*}

In order to obtain the prediction scores for a total of $C$ classes at a time step $t$, the outputs from last layer of the LSTM are sent to a softmax layer estimating probabilities as:
\begin{align*}
& prob_c=\frac{\exp({{\bf w}_c}^T{\bf h}^K_{t}+b_c)}{\sum_{c'\in C}\exp({{\bf w}_{c'}}^T{\bf h}^K_{t}+b_{c'})},
\end{align*}
where $prob_c$, ${\bf w}_c$ and $b_c$ are respectively the probability prediction, the corresponding weight vector and the bias term of the $c$-th class. Such an LSTM network can be trained with the Back-Propagation Through Time (BPTT) algorithm~\cite{graves2005framewise},  which ``unrolls'' the model into a feed forward neural net and back-propagates to determine the optimal weights.

As shown in Figure~\ref{fig:framework}, we adopt two LSTM models for temporal modeling. With the two-stream CNN pipeline for feature extraction, we have a spatial feature set  $({\bf x}_1^s,{\bf x}_2^s,\ldots,{\bf x}_T^s)$ and a motion feature set $({\bf x}_1^m,{\bf x}_2^m,\ldots,{\bf x}_T^m)$. The learning process leads to a set of predictions $({\bf y}_1^s,{\bf y}_2^s,\ldots,{\bf y}_T^s)$ for the spatial part and another set $({\bf y}_1^m,{\bf y}_2^m,\ldots,{\bf y}_T^m)$ for the motion part. For both LSTM models, we adopt the last step output ${\bf y}_T$ as the video-level prediction scores, since the outputs at the last time step are based on the consideration of the entire sequence. We empirically observe that the last step outputs are better than pooling predictions from all the time steps.

\subsection{Regularized Feature Fusion Network}
Given both the spatial and the motion features, it is easy to understand that correlations may exist between them since both are computed on the same video (e.g., person-related static visual features and body motions). A good feature fusion method is supposed to be able to take advantages of the correlations, while also can maintain the unique characteristics to produce a better fused representation. In order to explore this important problem rather than using the simple late fusion as~\cite{DBLP:conf/nips/SimonyanZ14}, we employ a regularized feature fusion neural network, as shown in the middle part of Figure~\ref{fig:framework}. First, average pooling is adopted to aggregate the frame-level CNN features into video-level representations, which are used as the inputs of the fusion network. The input features are non-linearly mapped to another layer and then fused in a feature fusion layer, where we apply regularizations in the learning of the network weights.

Denote $N$ as the total number of training videos with both the spatial and the motion representations. For the $n$-th sample, it can be written as a 3-tuple as $({\bf x}^s_n, {\bf x}^m_n, {\bf y}_n)$, where ${\bf x}^s_n = \sum_{t=1}^T {{\bf x}^s_{n,t}}\in \mathbb{R}^{d_s}$ and ${\bf x}^m_n = \sum_{t=1}^T {{\bf x}^m_{n,t}} \in \mathbb{R}^{d_m}$ represent the averaged spatial and motion features respectively, and ${\bf y}_n$ is the corresponding label of the $n$-th sample. 

For the ease of discussion, let us consider a degenerated case first, where only one feature is available. Denote $g(\cdot)$ as the mapping of the neural network from inputs to outputs. The objective of network training is to minimize the following empirical loss:
\begin{equation}
\label{eq:nn}
    \min_{{\bf W}} \sum_{i=1}^{N} \|  g({\bf x}_{i})-{\bf y}_i \|^2 + \lambda_1 \Phi({\bf W}),
\end{equation}
where the first term measures the discrepancy between the output $g({\bf x}_{i})$ and the ground-truth label ${\bf y}_i$, and the second term is usually a Frobenius norm based regularizer to prevent over-fitting.

We now move on to discuss the case of fusion and prediction with two features. Note that the approach can be easily extended to support more than two input features. Specifically, we use a fusion layer (see Figure~\ref{fig:framework}) to absorb the spatial and temporal features into a fused representation. To exploit the correlations of the features, we regularize the fusion process with a structural $\ell_{21}$ norm, which is defined as $\|{\bf W}\|_{2,1}$ $=\sum_i \sqrt{\sum_j w_{ij}^2}$. Incorporating the $\ell_{21}$ norm in the standard deep neural network formulation, we arrive at the following optimization problem:
\begin{equation}
\label{eq:obj1}
  \min_{{\bf W}} \mathcal{L} + \lambda_1 \Phi({\bf W})
+\frac{\lambda_{2}}{2}\left \| {\bf W}^{E} \right \|_{2,1},
\end{equation}
where  $\mathcal{L} =  \sum_{i=1}^{N} \|  g({\bf x}_{i}^{s},{\bf x}_{i}^{m})-{\bf y}_i \|^2$,  ${\bf W}^E = [{\bf W}^{E}_{s}, {\bf W}^{E}_{m}] \in \mathbb{R}^{P \times D}$ denotes the concatenated weight matrix from the $E$th layer (i.e., the last layer of feature abstraction in Figure~\ref{fig:framework}), $D = d_s + d_m$ and $P$ is the dimension of the fusion layer. 

Different from the objective in Equation~(\ref{eq:nn}), here we have an additional $\ell_{21}$ norm that is used for exploring feature correlations in the $E$-th layer. The term $\|{\bf W}^E\|_{2,1}$ computes the 2-norm of the weight values across different features in each dimension. Therefore, the regularization attains minimum when $\bf{W}^E$ contains the smallest number of non-zero rows, which is the discriminative information shared by distinct features. That is to say, the $\ell_{21}$ norm encourages the matrix ${\bf W}^E$ to be row sparse, which leads to similar zero/nonzero patterns of the columns of the matrix ${\bf W}^E$. Hence it enforces different features to share a subset of hidden neurons, reflecting the feature correlations.

However, in addition to seeking for the correlations shared among features, the unique discriminative information should also be preserved so that the complementary information can be used for improved classification performance. Thus, we add an additional regularizer to Equation~(\ref{eq:obj1}) as following:
\begin{equation}
\label{eq:obj}
  \min_{{\bf W}} \mathcal{L} + \lambda_1 \Phi({\bf W}) +\frac{\lambda_{2}}{2}\left \| {\bf W}^{E} \right \|_{2,1} + \lambda_{3}\left \| {\bf W}^{E} \right \|_{1,1}.
\end{equation}
The term $\|{\bf W}^E\|_{1,1}$ can be regarded as a complement of the $\|{\bf W}^E\|_{2,1}$ norm. It provides the robustness of the $\ell_{21}$ norm from sharing incorrect features among different representations, and thus allows different representations to emphasize different hidden neurons. 

Although the regularizer terms in Equation~(\ref{eq:obj}) are all convex functions, the optimization problem in Equation~(\ref{eq:obj}) is nonconvex due to the nonconvexity of the sigmoid function. Below, we discuss the optimization strategy using the gradient descent method in two cases:
\begin{enumerate}
    \item For the $E$-th layer, our objective function has four valid terms: the empirical loss, the $\ell_2$ regularizer $\Phi({\bf W})$, and two nonsomooth structural regularizers, \textit{i.e.}, the $\ell_{21}$ and $\ell_{11}$ terms. Note that simply using the gradient descent method is not optimal due to the two nonsmooth terms. We propose to optimize the $E$-th layer using a proximal gradient descent method, which splits the objective function into two parts:
        \begin{align*}
            p &= \mathcal{L} +  \lambda_1 \Phi({\bf W}), \\
            q &= \frac{\lambda_{2}}{2}\left \| {\bf W}^{E} \right \|_{2,1}+\lambda_{3}\left \| {\bf W}^{E} \right \|_{1,1},
        \end{align*}
        where $p$ is a smooth function and $q$ is a nonsmooth function. Thus, the update of the $i$-th iteration is formulated as:
        \begin{equation*}
            ({\bf W}^E)^{(i)} = \text{Prox}_q (({\bf W}^E)^{(i)} - \nabla p(({\bf W}^E)^{(i)})),
        \end{equation*}
        where $\text{Prox}$ is a proximal operator defined as:
        \begin{equation*}
            \text{Prox}_q({\bf W}) = \arg\min_{\bf V} \|{\bf W}-{\bf V}\| + q(V).
            \label{eq:prox}
        \end{equation*}
        The proximal operator on the combination of $\ell_{21}/\ell_{11}$ norm ball can be computed analytically as:
        \begin{equation}
        \label{eq:sol}
            {\bf W}^E_{r\cdot} = \left( 1- \frac{\lambda_2}{\|{\bf U}_{r\cdot}\|_2} \right) {\bf U}_{r\cdot}, \forall r = 1,\cdots, P,
        \end{equation}
        where ${\bf U}_{r\cdot} = \left[ |{\bf V}_{r\cdot}| - \lambda_3 \right]_{+} \cdot sign[{\bf V}_{r\cdot}]$, and ${\bf W}_{r\cdot},{\bf U}_{r\cdot},{\bf V}_{r\cdot}$ denote the $r$-th row of matrix ${\bf W}, {\bf U}$ and ${\bf V}$, respectively. Readers may refer to \cite{YangTKDD} for a detailed proof of a similar analytical solution.
    \item For all the other layers, the objective function in Equation~\ref{eq:obj} only contains the first two valid terms, which are both smooth. Thus, we can directly apply the gradient descent method as in \cite{bengio2012practical}. Denote ${\bf G}^l$ as the gradient of ${\bf W}^l$, the weight matrix of the $l$th layer is updated as:
    \begin{equation}
    \label{eq:update}
    {\bf W}^l = {\bf W}^l - \eta {\bf G}^l.
    \end{equation}
\end{enumerate}

The only additional computation cost for training our regularized feature fusion network is to compute the proximal operator in the $E$-th layer. The complexity of the analytical solution in Equation~(\ref{eq:sol}) is $O(P \times D)$. Therefore, the proposed proximal gradient descent method can quickly train the network with affordable computational cost. When incorporating more features, our formulation can be computed efficiently with linearly increased cost, while cubic operations are required by the approach of \cite{mm14:videoclassification} to reach a similar goal. In sum, the above optimization is incorporated into the conventional back propagation procedure, as described in Algorithm~\ref{alg}. 

\IncMargin{0.66em}
\begin{algorithm}
\SetKwData{Left}{left}\SetKwData{This}{this}\SetKwData{Up}{up}
\SetKwFunction{Union}{Union}\SetKwFunction{FindCompress}{FindCompress}
\SetKwInOut{Input}{Input}\SetKwInOut{Output}{Output}
\Input{$\mathbf{x}_{n}^s$ and $\mathbf{x}_{n}^m$: the spatial and motion CNN features of the $n$-th video sample\;
~~~~~~~~~~~~~$\mathbf{y}_n$: the label of the $n$-th video sample\;
~~~~~~~~~~~~~randomly initialized weight matrices ${\bf W}$;}
\Begin{
\For{$epoch\leftarrow 1$ \KwTo $M$}{
 Get the prediction error with feedforward propagation\;
 \For{$l \leftarrow L$ \KwTo 1} {
   Evaluate the gradients and update the weight matrices using Equation~(\ref{eq:update})\;
   \If{~$l==E$~}{
   Evaluate the proximal operator according to Equation~(\ref{eq:sol})\;
   }}
}}
\caption{The training procedure of the regularized feature fusion network.}\label{alg}
\end{algorithm}\DecMargin{0em}

\subsection{Discussion}
The approach described above has the capability of modeling static spatial, short-term motion and long-term temporal clues, which are all very important in video content analysis. One may have noticed that the proposed hybrid deep learning framework contains several components that are separately trained. Joint training is feasible but not adopted in this current framework for the following reason. The joint training process is more complex and existing works exploring this aspect indicate that the performance gain is not very significant. In~\cite{DBLP:journals/corr/DonahueHGRVSD14}, the authors jointly trained the LSTM with a CNN for feature extraction, which only improved the performance on a benchmark dataset from 70.5\% to 71.1\%. Besides, an advantage of separate training is that the framework is more flexible, where a component can be replaced easily without the need of re-training the entire framework. For instance, more discriminative CNN models like the GoogLeNet~\cite{Szegedy:2014tb} and deeper RNN models~\cite{DBLP:journals/corr/ChungGCB15} can be used to replace the CNN and LSTM parts respectively.

In addition, as mentioned in Section 1, there could be alternative frameworks or models with similar capabilities. The main contribution of this work is to show that such a hybrid framework is very suitable for video classification. In addition to showing the effectiveness of the LSTM and the regularized feature fusion network, we also show that the combination of both in the hybrid framework can lead to significant improvements, particularly for long videos that contain rich temporal clues.

\section{Experiments}
\subsection{Experimental Setup}
\subsubsection{Datasets}

We adopt two popular datasets to evaluate the proposed hybrid deep learning framework. 

\textbf{UCF-101}~\cite{ucf101}. The UCF-101 dataset is one of the most popular action recognition benchmarks. It consists of 13,320 video clips of 101 human actions (27 hours in total). The 101 classes are divided into five groups: Body-Motion, Human-Human Interactions, Human-Object Interactions, Playing Musical Instruments and Sports. Following~\cite{THUMOS14}, we conduct evaluations using 3 train/test splits, which is currently the most popular setting in using this dataset. Results are measured by classification accuracy on each split and we report the mean accuracy over the three splits.

\textbf{Columbia Consumer Videos (CCV)}~\cite{icmr11:consumervideo}. The CCV dataset contains 9,317 YouTube videos annotated according to 20 classes, which are mainly events like ``basketball'', ``graduation ceremony'', ``birthday party'' and ``parade''. We follow the convention defined in~\cite{icmr11:consumervideo} to use a training set of 4,659 videos and a test set of 4,658 videos. The performance is evaluated by average precision (AP) for each class, and we report the mean AP (mAP) as the overall measure.

\subsubsection{Implementation Details}
For feature extraction network structure, we adopt the VGG\_19~\cite{simonyan2014very} and the CNN\_M~\cite{DBLP:conf/nips/SimonyanZ14} to extract the spatial and the motion CNN features, respectively. The two networks can achieve 7.5\%~\cite{simonyan2014very} and 13.5\%~\cite{DBLP:conf/nips/SimonyanZ14} top-5 error rates on the ImageNet ILSVRC-2012 validation set respectively. The spatial CNN is first pre-trained with the ILSVRC-2012 training set with 1.2 million images and then fine-tuned using the video data, which is observed to be better than training from scratch. The motion CNN is trained from scratch as there is no off-the-shelf training set in the required form. In addition, simple data augmentation methods like cropping and flipping are utilized following~\cite{DBLP:conf/nips/SimonyanZ14}. 

The CNN models are trained using mini-batch stochastic gradient descent with a momentum fixed to 0.9. In the fine-tuning case of the spatial CNN, the rate starts from $10^{-3}$ and decreases to $10^{-4}$ after 14K iterations, then to $10^{-5}$ after 20K iterations. This setting is similar to~\cite{DBLP:conf/nips/SimonyanZ14}, but we start from a smaller rate of $10^{-3}$ instead of $10^{-2}$. For the motion CNN, we set the learning rate to $10^{-2}$ initially, and reduce it to $10^{-3}$ after 100K iterations, then to $10^{-4}$ after 200K iterations. Our implementation is based on the publicly available Caffe toolbox~\cite{jia2014caffe} with modifications to support parallel training with multiple GPUs in a server.

For temporal modeling, we adopt two layers in the LSTM for both the spatial and the motion features. Each LSTM has 1,024 hidden units in the bottom layer and 512 hidden units in the other layer. The network weights are learnt using a parallel implementation of the BPTT algorithm with a mini-batch size of 10. In addition, the learning rate and momentum are set to $10^{-4}$ and 0.9 respectively. The training is stopped after 150K iterations for both datasets.

For the regularized feature fusion network, we use four layers of neurons as illustrated in the middle of Figure~\ref{fig:framework}. Specifically, we first use one layer with 200 neurons for the spatial and motion feature to perform feature abstraction separately, and then one layer with 200 neurons for feature fusion with the proposed regularized structural norms. Finally, the fused features are used to build a logistic regression model in the last layer for video classification. We set the learning rate to 0.7 and fix $\lambda_{1}$ to $3 \times 10^{-5} $ in order to prevent over-fitting. In addition, we tune $\lambda_{2}$ and $\lambda_{3}$ in the same range as $\lambda_{1}$ using cross-validation.

\subsubsection{Compared Approaches}

To validate the effectiveness of our approach, we compare with the following baseline or alternative methods: (1) \textbf{Two-stream CNN}. Our implementation produces similar overall results with the original work~\cite{DBLP:conf/nips/SimonyanZ14}. We also report the results of the individual spatial-stream and motion-stream CNN models, namely \textbf{Spatial CNN} and \textbf{Motion CNN}, respectively; (2) \textbf{Spatial LSTM}, which refers to the LSTM trained with the spatial CNN features; (3) \textbf{Motion LSTM}, the LSTM trained with the motion CNN features; (4) \textbf{SVM-based Early Fusion (SVM-EF)}. $\chi^2$-kernel is computed for each video-level CNN feature and then the two kernels are averged for classification; (5) \textbf{SVM-based Late Fusion (SVM-LF)}. Separate SVM classifiers are trained for each video-level CNN feature and the prediction outputs are averaged; (6) \textbf{Multiple Kernel Learning (SVM-MKL)}, which combines the two features with the $\ell_p$-norm MKL~\cite{kloft2011lp} by fixing $p=2$; 
(7) \textbf{Early Fusion with Neural Networks (NN-EF)}, which concatenates the two features into a long vector and then use a neural network for classification; (8) \textbf{Late Fusion with Neural Networks (NN-LF)}, which deploys a separate neural network for each feature and then uses the average output scores as the final prediction; (9) \textbf{Multimodal Deep Boltzmann Machines (M-DBM)}~\cite{ngiam2011multimodal,srivastava2012multimodal}, where feature fusion is performed using a neural network in a \emph{free} manner without regularizations; (10) \textbf{RDNN} \cite{mm14:videoclassification}, which also imposes regularizations in a neural network for feature fusion, using a formulation that has a much higher complexity than our approach.

The first three methods are a part of the proposed framework, which are evaluated as baselines to better understand the contribution of each individual component. The last seven methods focus on fusing the spatial and the motion features (outputs of the first fully-connected layer of the CNN models) for improved classification performance. We compare our regularized fusion network with all the seven methods.

\subsection{Results and Discussions}
\subsubsection{Temporal Modeling}

We first evaluate the LSTM to investigate the significance of leveraging the long-term temporal clues for video classification. The results and comparisons are summarized in Table~\ref{tbl:lstm}. The upper two groups in the table compare the LSTM models with the two-stream CNN, which performs classification by pooling video-level representations without considering the temporal order of the frames. On UCF-101, the Spatial LSTM is better than the spatial stream CNN, while the result of the Motion LSTM is slightly lower than that of the motion stream CNN. It is interesting to see that, on the spatial stream, the LSTM is even better than the state-of-the-art CNN, indicating that the temporal information is very important for human action modeling, which is fully discarded in the spatial stream CNN. Since the mechanism of the LSTM is totally different, these results are fairly appealing because it is potentially very complementary to the video-level classification based on feature pooling.

\begin{figure*}[t]
\centering
\epsfig{file=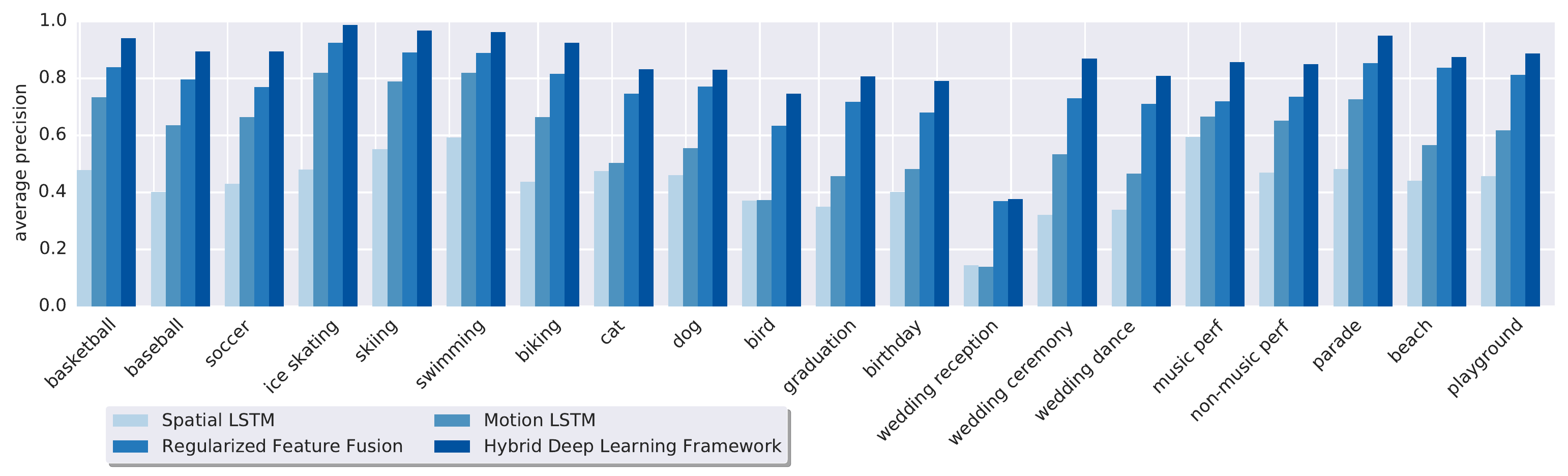, scale=0.57}
\vspace{-0.25in}
\caption{Per-class performance on CCV, using the Spatial and Motion LSTM, the Regularized Fusion Network, and their combination, i.e., the Hybrid Deep Learning Framework. }
\label{fig:perclassCCV}
\end{figure*}

On the CCV dataset, the LSTM models produce lower performance than the CNN models on both streams. The reasons are two-fold. First, since the average duration of the CCV videos (80 seconds) is around 10 times longer than that of the UCF-101 and the contents in CCV are more complex and noisy, the LSTM might be affected by the noisy video segments that are irrelevant to the major class of a video. Second, some classes like ``wedding reception" and ``beach" do not contain clear temporal order information (see Figure~\ref{fig:perclassCCV}), for which the LSTM can hardly capture helpful clues.

\begin{table}[h]
\begin{center}
\begin{tabular}{|c||c|c|}
\hline
              		& UCF-101 			& CCV    	\\ \hline  	\hline          
Spatial CNN   				& 80.1\% 			& 75.0\% 	\\ 
Spatial LSTM  				& 83.3\% 			& 43.3\% 	\\ \hline 	\hline
Motion CNN    				& 77.5\%			& 58.9\%    	\\ 
Motion LSTM   				& 76.6\%          	& 54.7\%    	\\ \hline 	\hline
CNN + LSTM (Spatial)	 	& 84.0\%            & 77.9\%     \\ \
CNN + LSTM (Motion)    		& 81.4\%			& 70.9\%     \\ 
CNN + LSTM (Spatial \& Motion)               
							& {\bf 90.1\%}            & {\bf 81.7\%}		\\ \hline
\end{tabular}
\caption{Performance of the LSTM and the CNN models trained with the spatial and the short-term motion features respectively on UCF-101 and CCV. ``+" indicates model fusion, which simply uses the average prediction scores of different models.}
\label{tbl:lstm} 
\end{center}
\end{table}

We now assess the performance of combining the LSTM and the CNN models to study whether they are complementary, by fusing the outputs of  the two types of models trained on the two streams. Note that the fusion method adopted here is the simple late average fusion, which uses the average prediction scores of different models. More advanced fusion methods will be evaluated in the next subsection. 

Results are reported in the bottom three rows of Table~\ref{tbl:lstm}. We observe significant improvements from model fusion on both datasets. On UCF-101, the fusion leads to an absolute performance gain of around 1\% compared with the best single model for the spatial stream, and a gain of 4\% for the motion stream.  On CCV, the improvements are more significant, especially for the motion stream where an absolute gain of 12\% is observed.  These results confirm the fact that the long-term temporal clues are highly complementary to the spatial and the short-term motion features. In addition, the fusion of all the CNN and the LSTM models trained on the two streams attains the highest performance on both datasets: 90.1\% and 81.7\% on UCF-101 and CCV respectively, showing that the spatial and the short-term motion features are also very complementary. Therefore, it is important to incorporate all of them into a successful video classification system. 

\begin{table}[t]
\begin{center}
\begin{tabular}{|c||c|c|}
\hline
		        & UCF-101      & CCV                              \\ \hline\hline

Spatial SVM		& 78.6\%          & 74.4\%                          \\ 
Motion SVM      & 78.2\%          & 57.9\%                          \\ 
\hline\hline

SVM-EF          	& 86.6\%          & 75.3\%                          \\ 
SVM-LF          	& 85.3\%          & 74.9\%                          \\  
SVM-MKL       		& 86.8\%          & 75.4\%                          \\ \hline  \hline
NN-EF           	& 86.5\%          & 75.6\%                          \\ 
NN-LF           	& 85.1\%          & 75.2\%                          \\ 
M-DBM          		& 86.9\%          & 75.3\%                          \\ 
Two-stream CNN		& 86.2\%  		  & 75.8\%							\\
RDNN 		& 88.1\%  		  & 75.9\%							\\ \hline \hline

Non-regularized Fusion Network   				
					& 87.0\% 		  &	75.4\%							\\ 
Regularized Fusion Network
					& {\bf 88.4\% }   & {\bf 76.2\% }             \\ \hline
\end{tabular}
\caption{\label{tbl:fusion} Performance comparison on UCF-101 and CCV, using various fusion approaches to combine the spatial and the short-term motion clues. }
\end{center}
\end{table}

\subsubsection{Feature Fusion}
Next, we compare our regularized feature fusion network with the alternative fusion methods, using both the spatial and the motion CNN features. The results are presented in Table~\ref{tbl:fusion}, which are divided into four groups. The first group reports the performance of individual features extracted from the first fully connected layer of the CNN models, classified by SVM classifiers. This is reported to study the gain from the SVM based fusion methods, as shown in the second group of results. The individual feature results using the CNN, i.e., the Spatial CNN and the Motion CNN, are already reported in Table~\ref{tbl:lstm}. The third group of results in Table~\ref{tbl:fusion} are based on the alternative neural network fusion methods. Note that NN-EF and NN-LF take the features from the CNN models and perform fusion and classification using separate neural networks, while the two-stream CNN approach performs classification using the CNN directly with a late score fusion (Figure 2). Finally, the last group contains the results of the proposed fusion network.

As can be seen, the SVM based fusion methods can greatly improve the results on UCF-101. On CCV, the gain is consistent but not very significant, indicating that the short-term motion is more important for modeling the human actions with clearer motion patterns and less noises. SVM-MKL is only slightly better than the simple early and late fusion methods, which is consistent with observations in recent works on visual recognition \cite{vedaldi2009multiple}.

Our proposed regularized feature fusion network (the last row in Table~\ref{tbl:fusion}) is consistently better than the alternative neural network based fusion methods shown in the third group of the table. In particular, the gap between our results and that of the M-DBM and the two-stream CNN confirms that using regularizations in the fusion process is helpful. Compared with the RDNN, our formulation produces slightly better results but with a much lower complexity as discussed earlier.

In addition, as the proposed formulation contains two structural norms, to directly evaluate the contribution of the norms, we also report a baseline using the same network structure without regularization (``non-regularized fusion network" in the table), which is similar to the M-DBM approach in its design but differs slightly in network structures. We see that adding regularizations in the same network can improve 1.4\% on UCF-101 and 0.8\% on CCV.

\subsubsection{The Hybrid Framework}
Finally, we discuss the results of the entire hybrid deep learning framework by further combining results from the temporal LSTM and the regularized fusion network. The prediction scores from these networks are fused linearly with weights estimated by cross-validation. As shown in the last row of Table~\ref{tb:comparison}, we achieve very strong performance on both datasets: 91.3\% on UCF-101 and 83.5\% on CCV. The performance of the hybrid framework is clearly better than that of the Spatial LSTM and the Motion LSTM (in Table~\ref{tbl:lstm}). Compared with the Regularized Fusion Network (in Table~\ref{tbl:fusion}), adding the long-term temporal modeling in the hybrid framework improves 2.9\% on UCF-101 and 7.3\% on CCV, which are fairly significant considering the difficulties of the two datasets. In contrast to the fusion result in the last row of Table~\ref{tbl:lstm}, the gain of the proposed hybrid framework comes from the use of the regularized fusion, which again verifies the effectiveness of our fusion method.

\begin{figure}[t!]
\centering
\epsfig{file=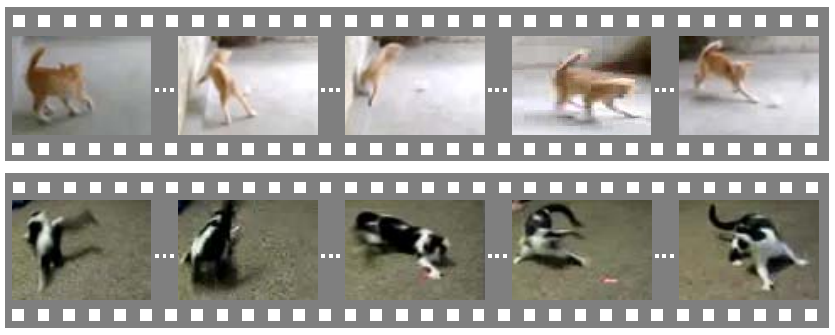, scale=1}
\vspace{-0.25in}
\caption{Two example videos of class ``cat" in the CCV dataset with similar temporal clues over time.}
\label{fig:example}
\end{figure}

To better understand the contributions of the key components in the hybrid framework, we further report the per-class performance on CCV in Figure~\ref{fig:perclassCCV}. We see that, although the performance of the LSTM is clearly lower, fusing it with the video-level predictions by the regularized fusion network can significantly improve the results for almost all the classes. This is a bit surprising because some classes do not seem to require temporal information to be recognized. After checking into some of the videos we find that there could be helpful clues which can be modeled, even for object-related classes like ``cat" and ``dog". For instance, as shown in Figure~\ref{fig:example}, we observe that quite a number of ``cat" videos contain only a single cat running around on the floor. The LSTM network may be able to capture this clue, which is helpful for classification.

\textbf{Efficiency.} In addition to achieving the superior classification performance, our framework also enjoys high computational  efficiency. We summarize the average testing time of a UCF-101 video (duration: 8 seconds) as follows. The extraction of the frames and the optical flows takes 3.9 seconds, and computing the CNN-based spatial and short-term motion features requires 9 seconds. Prediction with the LSTM and the regularized fusion network needs 2.8 seconds. All these are evaluated on a single NVIDIA Telsa K40 GPU. 

\subsubsection{Comparison with State of the Arts}
In this subsection, we compare with several state-of-the-art results.  As shown in Table~\ref{tbl:comparison}, our hybrid deep learning framework produces the highest performance on both datasets.  On the UCF-101, many works with competitive results are based on the dense trajectories~\cite{wang2013action,zha2015exploiting}, while our approach fully relies on the deep learning techniques. Compared with the original result of the two-stream CNN in~\cite{DBLP:conf/nips/SimonyanZ14}, our framework is better with the additional fusion and temporal modeling functions, although it is built on our implementation of the CNN models which are slightly worse than that of~\cite{DBLP:conf/nips/SimonyanZ14} (our two-stream CNN result is 86.2\%). Note that a gain of even just 1\% on the widely adopted UCF-101 dataset is generally considered as a significant progress.  In addition, the recent works in~\cite{DBLP:journals/corr/DonahueHGRVSD14,DBLP:journals/corr/SrivastavaMS15} also adopted the  LSTM to explore the temporal clues for video classification and reported promising performance. However, our LSTM results are not directly comparable as the input features are extracted by different neural networks.

On the CCV dataset, all the recent approaches relied on the joint use of multiple features by developing new fusion methods~\cite{xu2013feature,ye2012robust,MVA:audiovisual,DBLP:journals/ijcv/MaY14,liu2013sample,mm14:videoclassification}. Our hybrid deep learning framework is significantly better than all of them.

\begin{table}[t!]
\begin{center}
\begin{tabular}{|c|c||c|c|} 
\hline
\multicolumn{2}{|c||}{UCF-101}                &    \multicolumn{2}{c|}{CCV}        \\ \hline \hline

Donahue \textit{et al.}~\cite{DBLP:journals/corr/DonahueHGRVSD14}
& 82.9\% & Xu \textit{et al.}~\cite{xu2013feature}    & 60.3\% \\ \hline
Srivastava \textit{et al.}~\cite{DBLP:journals/corr/SrivastavaMS15} 
&	84.3\% & Ye \textit{et al.}~\cite{ye2012robust}		& 64.0\% \\ \hline
Wang \textit{et al.}~\cite{wang2013action}        &        85.9\%   			& Jhuo \textit{et al.}~\cite{MVA:audiovisual}& 64.0\% \\ \hline
Tran \textit{et al.}~\cite{tran2014c3d}        &        86.7\% 	    &  Ma \textit{et al.}~\cite{DBLP:journals/ijcv/MaY14}   & 63.4\% \\ \hline
Simonyan \textit{et al.}~\cite{DBLP:conf/nips/SimonyanZ14}   &      88.0\%       & Liu \textit{et al.}~\cite{liu2013sample}  & 68.2\% \\ \hline
Lan \textit{et al.}~\cite{lan2014beyond} &      89.1\%                & Wu \textit{et al.}~\cite{mm14:videoclassification}   & 70.6\% \\ \hline 
Zha \textit{et al.}~\cite{zha2015exploiting}    &        89.6\%     & 	\multicolumn{2}{c|}{/}  \\ \hline \hline
Ours  &  {\bf 91.3}\% & Ours &  {\bf 83.5}\% \\ \hline
\end{tabular}
\caption{\label{tbl:comparison} Comparison with state-of-the-art results. Our approach produces to-date the best reported results on both datasets.}
\label{tb:comparison}
\end{center}
\vspace{-0.2in}
\end{table}

\section{Conclusions}
We have proposed a novel hybrid deep learning framework for video classification, which is able to model static visual features, short-term motion patterns and long-term temporal clues. In the framework, we first extract spatial and motion features with two CNNs trained on static frames and stacked optical flows respectively. The two types of features are used separately as inputs of the LSTM network for long-term temporal modeling. A regularized fusion network is also deployed to combine the two features on video-level for improved classification. Our hybrid deep learning framework integrating both the LSTM and the regularized fusion network produces very impressive performance on two widely adopted benchmark datasets. Results not only verify the effectiveness of the individual components of the framework, but also demonstrate that the frame-level temporal modeling and the video-level fusion and classification are highly complementary, and a big leap of performance can be attained by combining them. 

Although deep learning based approaches have been successful in addressing many problems, effective network architectures are urgently needed for modeling sequential data like the videos. Several researchers have recently explored this direction. However, compared with the progress on image classification, the achieved performance gain on video classification over the traditional hand-crafted features is much less significant. Our work in this paper represents one of the few studies showing very strong results. For future work, further improving the capability of modeling the temporal dimension of videos is of high priority. In addition, audio features which are known to be useful for video classification can be easily incorporated into our framework.

\scriptsize
\bibliographystyle{abbrv}
\bibliography{reference}

\begin{thebibliography}{10}

\bibitem{arandjelovic2012three}
R.~Arandjelovic and A.~Zisserman.
\newblock Three things everyone should know to improve object retrieval.
\newblock In {\em CVPR}, 2012.

\bibitem{bach2004multiple}
F.~R. Bach, G.~R. Lanckriet, and M.~I. Jordan.
\newblock Multiple kernel learning, conic duality, and the smo algorithm.
\newblock In {\em ICML}, 2004.

\bibitem{bengio2012practical}
Y.~Bengio.
\newblock Practical recommendations for gradient-based training of deep
  architectures.
\newblock In {\em Neural Networks: Tricks of the Trade}. Springer, 2012.

\bibitem{DBLP:journals/corr/ChungGCB15}
J.~Chung, {\c{C}}.~G{\"{u}}l{\c{c}}ehre, K.~Cho, and Y.~Bengio.
\newblock Gated feedback recurrent neural networks.
\newblock {\em CoRR}, 2015.

\bibitem{Dahl:2012dx}
G.~E. Dahl, D.~Yu, L.~Deng, and A.~Acero.
\newblock {Context-Dependent Pre-Trained Deep Neural Networks for
  Large-Vocabulary Speech Recognition}.
\newblock {\em IEEE TASLP}, 2012.

\bibitem{DBLP:journals/corr/DonahueHGRVSD14}
J.~Donahue, L.~A. Hendricks, S.~Guadarrama, M.~Rohrbach, S.~Venugopalan,
  K.~Saenko, and T.~Darrell.
\newblock Long-term recurrent convolutional networks for visual recognition and
  description.
\newblock {\em CoRR}, 2014.

\bibitem{girshick2014rcnn}
R.~Girshick, J.~Donahue, T.~Darrell, and J.~Malik.
\newblock Rich feature hierarchies for accurate object detection and semantic
  segmentation.
\newblock In {\em CVPR}, 2014.

\bibitem{DBLP:conf/icassp/GravesMH13}
A.~Graves, A.~Mohamed, and G.~E. Hinton.
\newblock Speech recognition with deep recurrent neural networks.
\newblock In {\em ICASSP}, 2013.

\bibitem{graves2005framewise}
A.~Graves and J.~Schmidhuber.
\newblock Framewise phoneme classification with bidirectional lstm and other
  neural network architectures.
\newblock {\em Neural Networks}, 2005.

\bibitem{Harris1988}
C.~Harris and M.~J. Stephens.
\newblock A combined corner and edge detector.
\newblock In {\em Alvey Vision Conference}, 1988.

\bibitem{JainTHUMOS14}
M.~Jain, J.~van Gemert, and C.~G.~M. Snoek.
\newblock University of amsterdam at thumos challenge 2014.
\newblock In {\em ECCV THUMOS Challenge Workshop}, 2014.

\bibitem{MVA:audiovisual}
I.-H. Jhuo, G.~Ye, S.~Gao, D.~Liu, Y.-G. Jiang, D.~T. Lee, and S.-F. Chang.
\newblock Discovering joint audio-visual codewords for video event detection.
\newblock {\em Machine Vision and Applications}, 2014.

\bibitem{DBLP:conf/icml/JiXYY10}
S.~Ji, W.~Xu, M.~Yang, and K.~Yu.
\newblock 3d convolutional neural networks for human action recognition.
\newblock In {\em ICML}, 2010.

\bibitem{jia2014caffe}
Y.~Jia, E.~Shelhamer, J.~Donahue, S.~Karayev, J.~Long, R.~Girshick,
  S.~Guadarrama, and T.~Darrell.
\newblock Caffe: Convolutional architecture for fast feature embedding.
\newblock In {\em ACM Multimedia}, 2014.

\bibitem{jiang2009short}
W.~Jiang, C.~Cotton, S.-F. Chang, D.~Ellis, and A.~Loui.
\newblock Short-term audio-visual atoms for generic video concept
  classification.
\newblock In {\em ACM Multimedia}, 2009.

\bibitem{THUMOS14}
Y.-G. Jiang, J.~Liu, A.~Roshan~Zamir, G.~Toderici, I.~Laptev, M.~Shah, and
  R.~Sukthankar.
\newblock {THUMOS} challenge: Action recognition with a large number of
  classes.
\newblock \url{http://crcv.ucf.edu/THUMOS14/}, 2014.

\bibitem{icmr11:consumervideo}
Y.-G. Jiang, G.~Ye, S.-F. Chang, D.~Ellis, and A.~C. Loui.
\newblock Consumer video understanding: A benchmark database and an evaluation
  of human and machine performance.
\newblock In {\em {ACM} ICMR}, 2011.

\bibitem{KarpathyCVPR14}
A.~Karpathy, G.~Toderici, S.~Shetty, T.~Leung, R.~Sukthankar, and L.~Fei-Fei.
\newblock Large-scale video classification with convolutional neural networks.
\newblock In {\em CVPR}, 2014.

\bibitem{klaser2008spatio}
A.~Klaser, M.~Marsza{\l}ek, and C.~Schmid.
\newblock A spatio-temporal descriptor based on 3d-gradients.
\newblock In {\em BMVC}, 2008.

\bibitem{kloft2011lp}
M.~Kloft, U.~Brefeld, S.~Sonnenburg, and A.~Zien.
\newblock Lp-norm multiple kernel learning.
\newblock {\em The Journal of Machine Learning Research}, 2011.

\bibitem{krizhevsky2012imagenet}
A.~Krizhevsky, I.~Sutskever, and G.~E. Hinton.
\newblock Imagenet classification with deep convolutional neural networks.
\newblock In {\em NIPS}, 2012.

\bibitem{lan2014beyond}
Z.~Lan, M.~Lin, X.~Li, A.~G. Hauptmann, and B.~Raj.
\newblock Beyond gaussian pyramid: Multi-skip feature stacking for action
  recognition.
\newblock {\em CoRR}, 2014.

\bibitem{laptevSTIP}
I.~Laptev.
\newblock On space-time interest points.
\newblock {\em IJCV}, 64(2/3):107--123, 2007.

\bibitem{li2008expandable}
W.~Li, Z.~Zhang, and Z.~Liu.
\newblock Expandable data-driven graphical modeling of human actions based on
  salient postures.
\newblock {\em IEEE TCSVT}, 2008.

\bibitem{liu2013sample}
D.~Liu, K.-T. Lai, G.~Ye, M.-S. Chen, and S.-F. Chang.
\newblock Sample-specific late fusion for visual category recognition.
\newblock In {\em CVPR}, 2013.

\bibitem{lowe2004distinctive}
D.~G. Lowe.
\newblock Distinctive image features from scale-invariant keypoints.
\newblock {\em IJCV}, 2004.

\bibitem{DBLP:journals/ijcv/MaY14}
A.~J. Ma and P.~C. Yuen.
\newblock Reduced analytic dependency modeling: Robust fusion for visual
  recognition.
\newblock {\em IJCV}, 2014.

\bibitem{ngiam2011multimodal}
J.~Ngiam, A.~Khosla, M.~Kim, J.~Nam, H.~Lee, and A.~Ng.
\newblock Multimodal deep learning.
\newblock In {\em ICML}, 2011.

\bibitem{oneata2013action}
D.~Oneata, J.~Verbeek, C.~Schmid, et~al.
\newblock Action and event recognition with fisher vectors on a compact feature
  set.
\newblock In {\em ICCV}, 2013.

\bibitem{DBLP:journals/corr/RanzatoSBMCC14}
M.~Ranzato, A.~Szlam, J.~Bruna, M.~Mathieu, R.~Collobert, and S.~Chopra.
\newblock Video (language) modeling: a baseline for generative models of
  natural videos.
\newblock {\em CoRR}, 2014.

\bibitem{DBLP:journals/corr/RazavianASC14}
A.~S. Razavian, H.~Azizpour, J.~Sullivan, and S.~Carlsson.
\newblock {CNN} features off-the-shelf: an astounding baseline for recognition.
\newblock {\em CoRR}, 2014.

\bibitem{sanchez2013image}
J.~S{\'a}nchez, F.~Perronnin, T.~Mensink, and J.~Verbeek.
\newblock Image classification with the fisher vector: Theory and practice.
\newblock {\em IJCV}, 2013.

\bibitem{DBLP:conf/nips/SimonyanZ14}
K.~Simonyan and A.~Zisserman.
\newblock Two-stream convolutional networks for action recognition in videos.
\newblock In {\em NIPS}, 2014.

\bibitem{simonyan2014very}
K.~Simonyan and A.~Zisserman.
\newblock Very deep convolutional networks for large-scale image recognition.
\newblock {\em CoRR}, 2014.

\bibitem{ucf101}
K.~Soomro, A.~R. Zamir, and M.~Shah.
\newblock {UCF101:} {A} dataset of 101 human actions classes from videos in the
  wild.
\newblock {\em CoRR}, 2012.

\bibitem{DBLP:journals/corr/SrivastavaMS15}
N.~Srivastava, E.~Mansimov, and R.~Salakhutdinov.
\newblock Unsupervised learning of video representations using {LSTMs}.
\newblock {\em CoRR}, 2015.

\bibitem{srivastava2012multimodal}
N.~Srivastava and R.~Salakhutdinov.
\newblock Multimodal learning with deep boltzmann machines.
\newblock In {\em NIPS}, 2012.

\bibitem{Szegedy:2014tb}
C.~Szegedy, W.~Liu, Y.~Jia, P.~Sermanet, S.~Reed, D.~Anguelov, D.~Erhan,
  V.~Vanhoucke, and A.~Rabinovich.
\newblock {Going Deeper with Convolutions}.
\newblock {\em CoRR}, 2014.

\bibitem{tang2012learning}
K.~Tang, L.~Fei-Fei, and D.~Koller.
\newblock Learning latent temporal structure for complex event detection.
\newblock In {\em CVPR}, 2012.

\bibitem{tran2014c3d}
D.~Tran, L.~Bourdev, R.~Fergus, L.~Torresani, and M.~Paluri.
\newblock C3d: Generic features for video analysis.
\newblock {\em CoRR}, 2014.

\bibitem{vail2007conditional}
D.~L. Vail, M.~M. Veloso, and J.~D. Lafferty.
\newblock Conditional random fields for activity recognition.
\newblock In {\em AAAMS}, 2007.

\bibitem{vedaldi2009multiple}
A.~Vedaldi, V.~Gulshan, M.~Varma, and A.~Zisserman.
\newblock Multiple kernels for object detection.
\newblock In {\em ICCV}, 2009.

\bibitem{DBLP:journals/corr/VenugopalanXDRMS14}
S.~Venugopalan, H.~Xu, J.~Donahue, M.~Rohrbach, R.~J. Mooney, and K.~Saenko.
\newblock Translating videos to natural language using deep recurrent neural
  networks.
\newblock {\em CoRR}, 2014.

\bibitem{wang2013action}
H.~Wang and C.~Schmid.
\newblock Action recognition with improved trajectories.
\newblock In {\em ICCV}, 2013.

\bibitem{wang2009evaluation}
H.~Wang, M.~M. Ullah, A.~Klaser, I.~Laptev, and C.~Schmid.
\newblock Evaluation of local spatio-temporal features for action recognition.
\newblock In {\em BMVC}, 2009.

\bibitem{wang2009max}
Y.~Wang and G.~Mori.
\newblock Max-margin hidden conditional random fields for human action
  recognition.
\newblock In {\em CVPR}, 2009.

\bibitem{mm14:videoclassification}
Z.~Wu, Y.-G. Jiang, J.~Wang, J.~Pu, and X.~Xue.
\newblock Exploring inter-feature and inter-class relationships with deep
  neural networks for video classification.
\newblock In {\em ACM Multimedia}, 2014.

\bibitem{xu2013feature}
Z.~Xu, Y.~Yang, I.~Tsang, N.~Sebe, and A.~Hauptmann.
\newblock Feature weighting via optimal thresholding for video analysis.
\newblock In {\em ICCV}, 2013.

\bibitem{YangTKDD}
H.~Yang, M.~R. Lyu, and I.~King.
\newblock Efficient online learning for multitask feature selection.
\newblock {\em ACM SIGKDD}, 2013.

\bibitem{ye2012robust}
G.~Ye, D.~Liu, I.-H. Jhuo, and S.-F. Chang.
\newblock Robust late fusion with rank minimization.
\newblock In {\em CVPR}, 2012.

\bibitem{zeng2010knowledge}
Z.~Zeng and Q.~Ji.
\newblock Knowledge based activity recognition with dynamic bayesian network.
\newblock In {\em ECCV}, 2010.

\bibitem{zha2015exploiting}
S.~Zha, F.~Luisier, W.~Andrews, N.~Srivastava, and R.~Salakhutdinov.
\newblock Exploiting image-trained cnn architectures for unconstrained video
  classification.
\newblock {\em CoRR}, 2015.

\bibitem{zhang2007local}
J.~Zhang, M.~Marsza{\l}ek, S.~Lazebnik, and C.~Schmid.
\newblock Local features and kernels for classification of texture and object
  categories: A comprehensive study.
\newblock {\em IJCV}, 2007.

\end{thebibliography}

\end{document}